# AI, Native Supercomputing and the Revival of Moore's Law

CHIEN-PING LU, NovuMind Inc

**Abstract**

Artificial Intelligence (AI) was the inspiration that shaped Computing as we know it today. In this article, I explore why and how AI would continue to inspire Computing and reinvent it when Moore's Law is running out of steam.

At the dawn of Computing, Alan Turing proposed that instead of comprising many different specific machines, the computing machinery to make machines think should be a universal digital computer, modeled after human *computers* carrying out calculations with pencil on paper. Based on the belief that a digital computer would be significantly faster, more diligent and patient than a human, he anticipated that AI would be advanced as *software*. Given the challenge to write all the necessary software to make machines think, he also envisioned a *learning machine* to generate the software. Nevertheless, the focus of the industry has been on the universal computers, which have become exponentially faster and more energy efficient through Moore's Law. Even though software has not yet made a machine think, it has been changing how we live fundamentally. The first Computing revolution started when the software was decoupled from the computing machinery.

Since the slowdown of Moore's Law in 2005, the universal computer is no longer improving exponentially in terms of speed and energy efficiency. It has to carry legacy, and cannot be aggressively modified to save energy. Turing's proposition of AI-as-software is challenged, and the temptation of making many domain-specific AI machines emerges. Thanks to Deep Learning, we need to build only one domain-specific machine, the universal learning machine. The corresponding software stays decoupled from the computing machinery in the language of linear algebra, which it has in common with Supercomputing. The new computing machinery for AI consists of a universal computer and a universal learning machine. The later understands linear algebra natively to then become a Native Supercomputer.

AI has been and will still be the inspiration for Computing. The quest to make machines think continues amid the slowdown of Moore's Law. AI might not only maximize the remaining benefits of Moore's Law, but also revive Moore's Law beyond current technology.

**Keywords** AI, Deep Learning, Moore's Law, History of Computing, Parallel Computing, Supercomputing, High Performance Computing, Dataflow

---

## AI and the Universal Computer

What kind of computing machinery do we need to advance AI to human level? At the dawn of Computing, one of the founding fathers, Alan Turing, believed that AI could be approached as *software* running on a universal computer. This was a revolutionary idea given that during his time, the term "computer" was generally referred to as a human hired to do calculations with pencil on paper. Turing referred to a machine as a "digital computer" to distinguish it from the human one.

In the context of AI, Alan Turing is remembered for his *Imitation Game*, or later referred to as *Turing Test*, in which a machine strives to exhibit intelligence to make itself indistinguishable from a human in the eyes of an interrogator. In his landmark paper, "Computing Machinery and Intelligence" (Turing, 1950), he tried to address the ultimate AI question, "Can machines think?" He reframed the question more precisely and unambiguously by asking how well a machine does in the imitation game. Turing hypothesized that human intelligence is "computable," which has a precise mathematical meaning famously established by himself (Turing, 1936), as a bag of discrete state machines, and reframed the ultimate AI question as

> Are there discrete machines that would do well **(in the imitation game)**? (Turing, 1950)

But what exactly are the discrete state machines to win the imitation game? Apparently, he did not know during his time. But witnessing the extreme difficulty of building a non-human, electronic computer himself (Turing, 1950b), he envisioned only one machine, the Universal Digital Computer that could mimic any discrete state machine. Each discrete state machine can be encoded as numbers to be processed by a universal computer. The numbers that encode a discrete state machine become software, and the computing machinery became the "stored program computer" envisioned by John von Neumann in his incomplete report (Neumann, 1945). Thus, Turing concluded:

> Considerations of speed apart, it is unnecessary to design various new machines to do various computing processes. They can all be done with one digital computer, suitably programmed for each case. (Turing, 1950)

Thereafter, the history of computing has been mainly the race to build faster Universal Computers to answer the following challenge:

> Are there imaginable digital computers that would do well **(in the imitation game)**? (Turing, 1950)

AI researchers and thinkers have been advancing AI without worrying about the underlying computing machinery. People might argue that this applies only to traditional rule-based AI. However, even connectionists have to translate their connectionist systems into algorithms in software to prove and demonstrate their ideas. We have been seeing advances and innovations in Deep Learning completely decoupled from the underlying computing machinery. Today, we use terms like "machines", "networks", "neurons", and "synapses", without a second thought about the fact that those entities do not have to exist physically. People ponder about a grand unified theory of Deep Learning using ideas like "emergent behaviors", "intuitions", "nonlinear dynamics", believing that those concepts could be adequately represented or approximated by software. AI has been and will be advanced as software.

---

**The Perfect Marriage between The Universal Computer and Moore's Law**

Turing's Universal Computer inspired von Neumann to come up with a powerful computing paradigm, in which complex functions were expressed in a simple yet complete language, the Instruction Set Architecture (ISA), that computing machinery could understand and execute. It brought us computers, as well as the software industry. The prevailing computing machinery in the era of von Neumann paradigm is the microprocessor, now a synonym of the Central Processing Unit (CPU), designed to run instructions in stored programs sequentially. The CPU, the Graphics Processing Unit (GPU), and the various kinds of Digital Signal Processor (DSP) and programmable alternatives are all modern incarnations of such a Universal Digital Computer.

But how would such a computer, emulating non-intelligent and non-thinking behaviors of a human, demonstrate human-level intelligence? Turing's answer was this:

> Provided it could be carried out sufficiently quickly the digital computer could mimic the behaviour of any discrete state machine. (Turing, 1950)

As far as AI is concerned, Turing's idea was that AI can fundamentally be approached through software running on a Universal Digital Computer. It would be the responsibility of the architects of the computing machinery to make it sufficiently fast. But how would we make it faster and at what rate?

Moore's Law, coined in Gordon Moore's seminal paper (Moore, 1965), has been followed by the semiconductor industry as a consensus and commitment to double the number of transistors per area every two years. Based on the technology scaling rule called Dennard Scaling, transistors have not only become smaller, but also faster and more energy efficient such that a chip now offers at least twice the performance at roughly the same dollar and power budgets. The performance growth mainly came from Moore's Law driving the clock speed exponentially faster. From 1982 to 2005, typical CPU clock speed grew by 500 times from 6 MHz to 3 GHz. Computing machinery vendors strived to build more capable CPUs, through faster clock speeds and capacity to do more than one thing at a time while maintaining the sequential semantics of a universal computer. Software vendors endeavored to explore new application scenarios and solve the problems algorithmically. The decoupling of software from the computing machinery and the scaling power of Moore's Law triggered the Computing revolution that has made today's smart phones more powerful than supercomputers two decades ago.

However, faster computers have not helped AI pass the Turing Test yet. AI started out as a discipline to model intelligence behaviors with algorithmic programs following the von Neumann paradigm. It had been struggling to solve real world problems and waiting for even faster computers. Unfortunately, the exponential performance growth of a universal computer has ground to a halt.

**The Slowdown of Moore's Law**

The turning point happened in 2005, when the transistors, while continuing to double in numbers, were neither faster nor more energy efficient at the same rates as before due to the breakdown of Dennard Scaling. Intel wasted little time to bury the race for faster clock speed, and introduced multi-core to keep up performance by running multiple "cores" in parallel. A universal computer became a "core" in a multi-core CPU, or a GPU. Multi-core has been a synonym of Parallel Computing in the CPU community. It was expected that there would be a smooth transition from von Neumann paradigm to its parallel heir, and the race for faster clock speed would be replaced with one for higher core count starting from dual and quad cores, to eventually a sea of cores. Around the same time, programmers were asked to take on the challenge of writing and managing a sea of programs, or "threads" (Sutter, 2005).

Such a race to double core count has not happened. Intel and the CPU industry have been struggling to add cores aggressively due to the issue of lagging improvements in transistor energy efficiency, manifested as the Dark Silicon phenomenon. It implies that while being able to accommodate four times more cores on a die through two generations of transistor shrinking, we could power up only half of the cores. If this does not look serious enough, only one quarter of the cores can be powered up at the third generation of transistor shrinking. Unless we reduce the core aggressively to compensate for the lagging improvement in energy efficiency, there might be no incentive to go with the fourth generation of transistor shrinking as there will be negligible performance improvement (see Figure 1). To make the situation even worse, the gap between the speed of memory and that of logic has been widening exponentially.

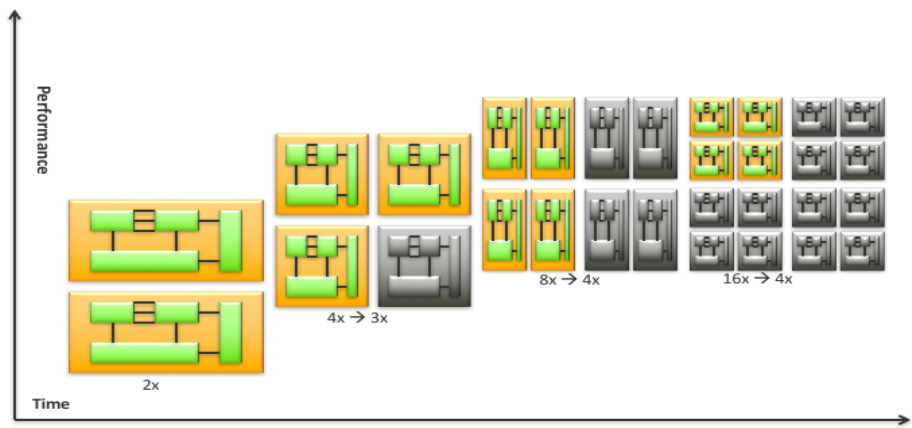

Figure 1 Dark Silicon phenomenon: diminishing returns with more cores

Such a limit applies to any computing machinery with an ISA legacy to carry, including the GPU. Although the GPU does not need to support ISA compatibility to every bit, it still needs to support higher level standards such as OpenGL and DirectX shading languages and OpenCL, and intermediate level standards such as SPIR-V. NVIDIA needs to maintain the legacy in their propriety CUDA. For software, managing the threads explicitly for a sea of cores has turned out to be untenable unless we restrict the communications among the threads to some patterns. *Such massive and unwieldy parallelism is not for the computing machinery and software to tackle*.

Some prominent research on Dark Silicon, such as "Dark Silicon and the End of Multicore Scaling" by Hadi Esmaeilzadeh (2011), confused the physical limitation in semiconductor with that from Amdahl's law, and prematurely declared the death of parallelism along with the slowdown of Moore's Law. There is abundant parallelism in AI with Deep Learning as we will see later. Once AI with Deep Learning replaces von Neumann architecture as the dominant computing paradigm, abundant parallelism will be the norm.

**Moore's Law and AI**

Turing was not specific about the performance and energy efficiency of a universal computer. He assumed that computers would always be sufficiently fast, and would not be a gating factor for the quest for human-level AI. But if passing the Turing Test is the ultimate criteria for machine intelligence, he would have suggested that the computers must achieve a certain level of performance and efficiency to exhibit intelligence; otherwise, the interrogator would be suspicious if it takes too long for a computer to respond to questions or consumes too many resources in the effort.

Turing envisioned his digital computer as one that models the slow thinking process of a human doing calculations with a pencil on a piece of paper. The universal digital computer was named to imply that it was designed to model after a human "computer." According to Turing:

> The human computer is supposed to be following fixed rules; he has no authority to deviate from them in any detail. (Turing, 1950)

In other words, such a universal digital computer does not think, but follows the instructions provided by software. It is the software that makes it think. Following fixed rules strictly requires intensive concentration and is an energy-consuming and slow process for a human brain. Try to multiply 123 by 456 in your head while you are running. It will slow you down. Interestingly, what's energy consuming for human is also for computers. To accomplish a task by executing one instruction at a time takes relatively more energy than doing it natively without the intermediate ISA. Approaching AI as software in the von Neumann paradigm is like mimicking fast and effortless human mental functions, such as intuition, with a machine that is based on the slow mental process of a human.

Turing did not foresee that a universal computer would run out of steam. If we are to stay with the von Neumann computing paradigm, we need to put an army of universal computers in a machine to continue the quest. These universal computers would have to communicate data and coordinate tasks among them. However, the slowdown of Moore's Law and the legacy of the von Neumann paradigm suggest that we will not able to supply sufficient energy to keep such an army growing in size. There needs to be a paradigm shift for AI and Computing.

Turing did foresee that it would be difficult for human programmers to write all the necessary software to make machines think. He suggested that we implement a learning machine modeled after a child's mind:

> Instead of trying to produce a programme to simulate the adult mind, why not rather try to produce one which simulates the child's? If this were then subjected to an appropriate course of education one would obtain the adult brain. (Turing, 1950)

This idea leads to Deep Learning. Although Turing did not predict the emergence of Deep Learning, he was aware of the approach with Neural Networks:

> It is generally thought that there was always an antagonism between programming and the 'connectionist' approach of neural networks. But Turing never expressed such a dichotomy, writing that both approaches should be tried. (Hodges, 2013)

Turing probably considered a learning machine to be simulated on a universal computer as software like any other discrete state machine. However, given that it plays an essential role to generate many other programs, it seems reasonable to build a dedicated learning machine. If Turing were alive today and witnessed the emergence of Deep Learning and the slowdown Moore's Law, he would have revised his proposition on the computing machinery for AI to make the learning machine part of the machinery to compliment the universal computer.

**Deep Learning and the Universal Learning Machine**

Deep Learning has been transforming and consolidating AI since it came to the center stage of Computing in 2012. With Deep Learning, the intelligence is not coded directly by programmers but acquired indirectly through Neural Networks, which are able to learn every continuous function as shown in "Multilayer feedforward networks are universal approximators" by (Kurt Hornik, 1989). The acquisition and manifestation of the intelligence can be formulated as computations dominated by a compact set of linear algebra primitives analogous to those defined in BLAS (Basic Linear Algebra Subprograms), the fundamental application programming interface used in Supercomputing and High Performance Computing (HPC). With Deep Learning, AI and Supercomputing effectively speak the same language with dialectical variances in numerical precisions, and minor differences in domain-specific requirements.

As mentioned earlier, the massive and unwieldy parallelism under the von Neumann paradigm is not for the computing machinery and software to tackle. On the other hand, the patterns of parallelism in Supercomputing are not unwieldy and can be summarized as Collective Communication (see Figure 2) as described in Frank Capello's "Communication Determinism in Parallel HPC Applications" (2010). Collective Communication has been proven to be scalable and manageable in large-scale distributed supercomputing systems.

Through Deep Learning, the child machine can potentially be liberated from the von Neumann architecture to handle linear algebra computations and collective communication natively and more efficiently.

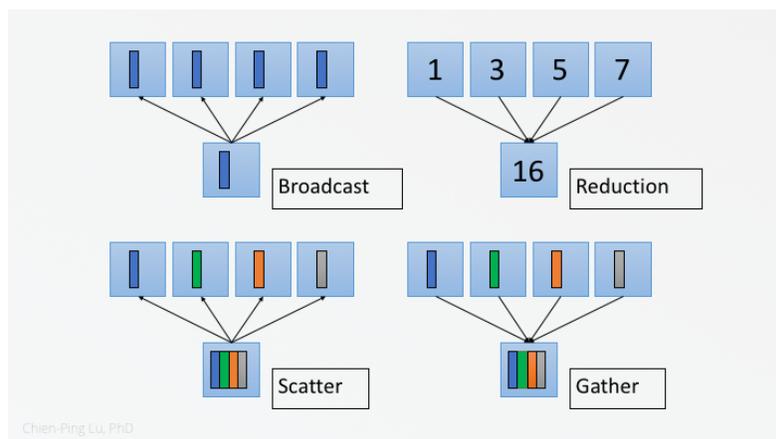

Figure 2 Four basic collective communication operations

**Why Linear Algebra?**

The fundamental primitives in Deep Learning are tensors, high-dimensional data arrays used to represent layers of Deep Neural Networks. A Deep Learning task can be described as a Tensor Computation Graph (Figure 3):

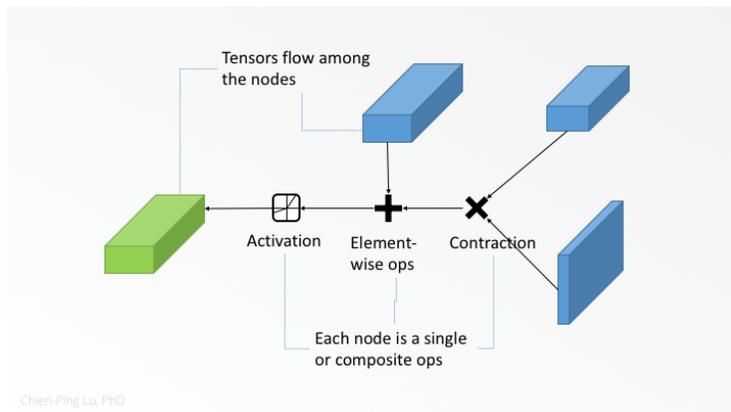

Figure 3 A tensor computation graph

A tensor computation graph is effectively a piece of AI software. A universal computer can "decode" such a graph, and instruct a child machine either to learn or to inference according to the graph. Tensors can be unfolded into 2-dimensional matrices, and matrix computations are handled thru matrix kernels (see Figure 4). Matrix kernels refer to CPU or GPU programs implementing different types of matrix computations comprising many MAC (multiply accumulate) operations. Such a matrix-centric approach is described in Sharan Chetlur (2014). The MAC operations for matrix multiplication are the most time-consuming part of Deep Learning. One might ask, if computations in Deep Learning are predominantly MACs in matrix computations, *why don't we simplify a core all the way to a MAC unit that does nothing but a MAC operation*? In fact, *why does a MAC unit need to keep the legacy of being a core at all*?

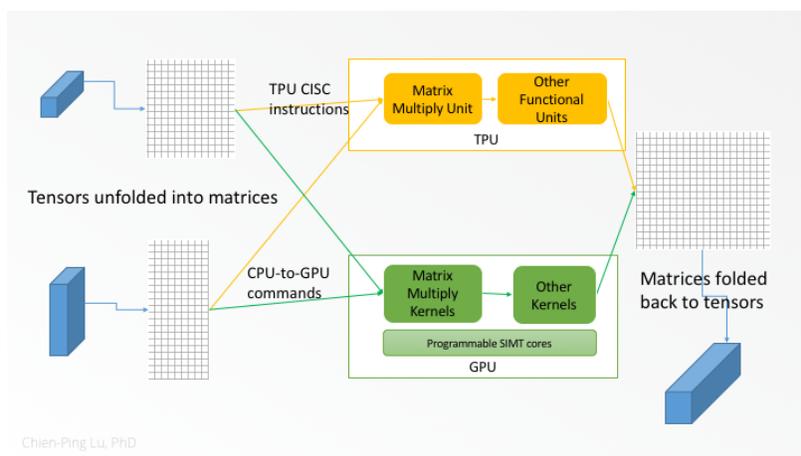

Figure 4 Matrix-centric platforms on the GPU and the TPU.

**The TPU and Systolic Arrays**

In the highly-anticipated paper, "In-Datacenter Performance Analysis of a Tensor Processing Unit" (Jouppi, 2017), Google disclosed the technical details and performance metrics of the Tensor Processing Unit (TPU). The TPU was built around a matrix multiply unit based on systolic arrays. What's eye-catching is the choice by the TPU design team to use a systolic array. A systolic array is a specific spatial dataflow machine. A Processing Element (PE) in a systolic array works in lock step with its neighbors. Each PE in a systolic array is basically a MAC unit with some glue logic to store and forward data. In comparison, a computing unit equivalent to a PE in a mesh-connected parallel processor is a full-featured processor core with its own frontend and necessary peripherals, whereas a PE equivalent in a GPU is a simplified processor core sharing a common frontend and peripherals with other cores in the same compute cluster. Among the three solutions, the density of MAC units is the highest in a systolic array. These differences are shown in Figure 5:

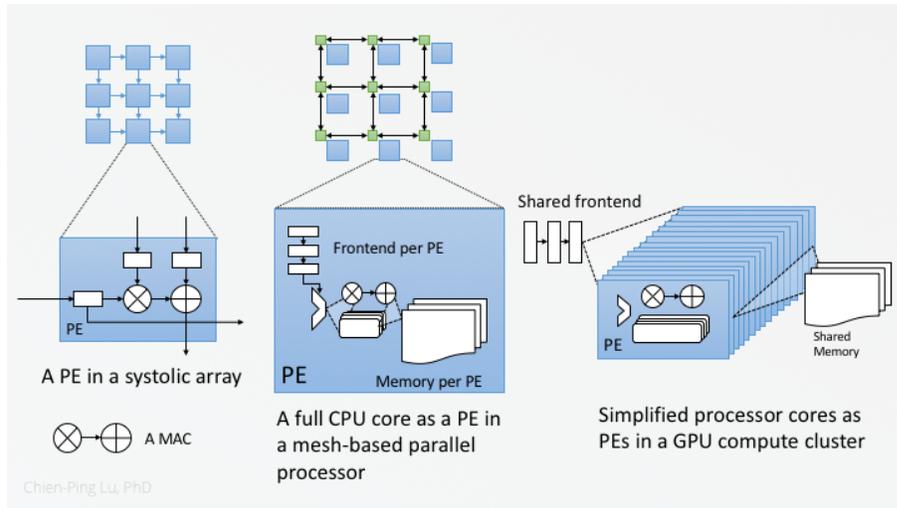

Figure 5 PEs in a systolic array, mesh-connected parallel processor and a GPU

A systolic array claims several advantages: simple and regular design, concurrency and communication, and balancing computation with I/O. However, until now, there has been no commercially successful processor based on a systolic array. The TPU is the first, and it is impressive, arguably the largest systolic array implemented or even conceived. Their design is reminiscent of an idea introduced by H. T. Kung (Kung, 1982). However, due to the curse of the square shape, it suffers from scalability issues as elaborated in the LinkedIn article, "Should We All Embrace Systolic Arrays" (Lu, 2017).

---

**Spatial Dataflow Architecture**

Like a systolic array, the building block of a generic spatial dataflow machine is often referred to as the PE, which is typically a MAC unit with some glue logic. Mesh topology is a strikingly popular way to organize PEs, for examples, Google's TPU (Jouppi, 2017), the DianNao family (Chen, 2014), MIT's Eyeriss (Sze, 2017). See Figure 6.

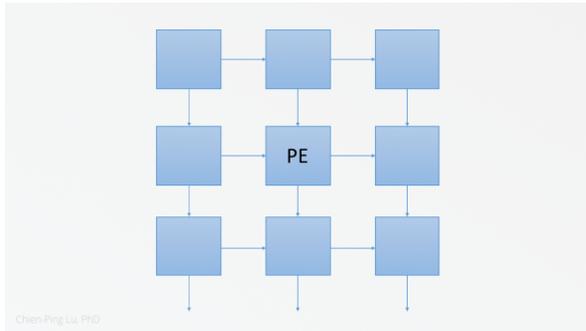
Figure 6 A PE and its neighbors in a mesh

It seems logical to use a mesh topology to organize the PEs on a 2-dimensional chip when there are lots of PEs and regularity is desirable. Such an arrangement leads to the following two *mesh-centric assumptions*:

1. The distance for a piece of data to travel across the mesh in one clock period is fixed as that between 2 neighboring PEs, even though it could be much further;
2. A PE depends on the upstream neighboring PEs to compute even though such a dependency mainly comes more from the spatial order, rather than from true data dependency.

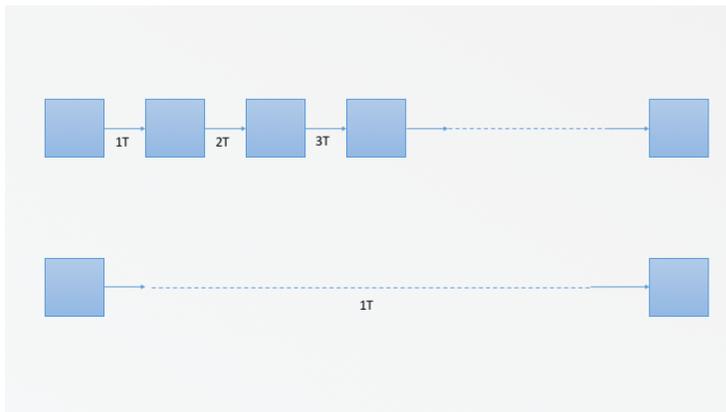
Figure 7 Mesh-centric assumption 1

The first assumption is a legacy inherited from distributed parallel processors comprising many compute nodes. Each compute node has to communicate among themselves through intermediate nodes. It is analogous to the situation when a high-speed train stops at every single station on the way to the destination, as shown in Figure 7. Within one clock period, a piece of data could travel over a distance equal to hundreds of the width of a MAC unit without having to hop over every single MAC unit in between. Restricting dataflows to PE hopping in a mesh topology causes an increase in latency by several orders of magnitude.

The second assumption is another legacy inherited from distributed parallel processors. Each compute node not only handles computations but also plays a part in the distributed storage of the data. The nodes need to exchange data among them to make forward progress. For an on-chip processing mesh, however, the data comes from the side interfacing with the memory. The data flow through the mesh and the results are collected on the other side as shown in Figure 8. Due to the local topology, an internal PE has to get the data through the PEs sitting between it and the memory. Likewise, it has to contribute its partial result through the intermediate PEs before reaching the memory. *The resulting dataflows are due to the spatial order of the PE in the mesh, not as a result of true data dependency.*

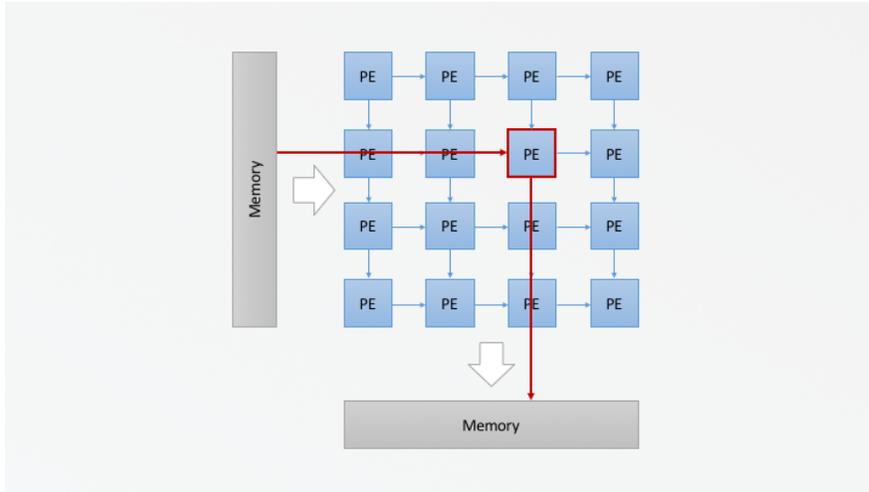

Figure 8 Mesh-centric assumption 2

Given the two mesh-centric assumptions, no matter how many PEs and how much bandwidth you have, the performance to solve a problem on a $d$-dimensional mesh is limited by the dimensionality $d$ of the mesh, not the number of the PEs, nor the IO bandwidth. Suppose a problem requires I inputs, $K$ outputs, and T computations, then the asymptotic running time to solve the problem on a d-dimensional mesh is given by Fisher's bound (Fisher, 1988):

$$t = \Omega(\max(\sqrt[d]{I}, \sqrt[d]{K}, \sqrt[d+1]{T})).$$

Fisher's bound implies there are upper bounds on the number of PEs and bandwidth beyond which no further running time improvement is achievable.

Applying Fisher's bound to the inner product, the running time to do an inner product is $\Omega(n)$ on a 1-dimensional mesh. If you can afford to have a 2-dimensional mesh, the running time is $\Omega(\sqrt{n})$. Can we do better? Instead of using 1 or 2-dimensional mesh, we can feed the input to $n$ PEs and add the products in pairs recursively. A $\Omega(\log(n))$ running time can be achieved. However, it is not possible to achieve such a performance on an either 1 or 2-dimensional mesh unless we organize the PEs in the way shown in Figure 9.

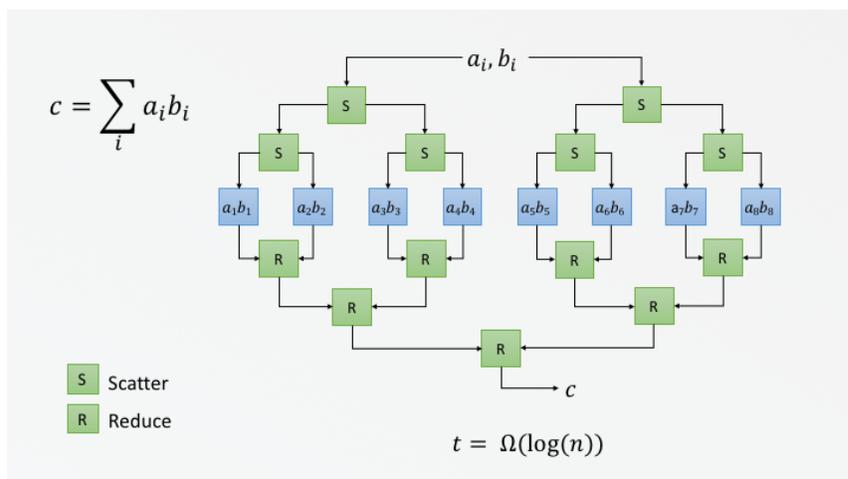

Figure 9 A faster inner products than Fisher's bound

The reasons for such a *super-optimal* result compared to the theoretical limits on a mesh is that there is no PE hopping, and it uses links of different lengths assuming that it takes the same time for a piece of data to travel over links with different lengths. If the distance is too long for a piece of data to travel in one clock period, we can add flops in the middle. It should be an implementation issue, not an architectural one.

**Matrix Multiplication According to Supercomputing**

Let's look at the most time-consuming part of Deep Learning: Matrix Multiplication, which has always been at the heart of Supercomputing. State-of-the-art parallel matrix multiplication performance on modern supercomputers is achieved with the following two major advancements:

1. Scalable matrix multiplication algorithms
2. Efficient collective communications with logarithmic overhead

**Scalable matrix multiplication algorithms**

See Figure 10 for the demonstration of matrix multiplication in outer products. The computations are 2-dimensional, but both the data and the communications among them are 1-dimensional.

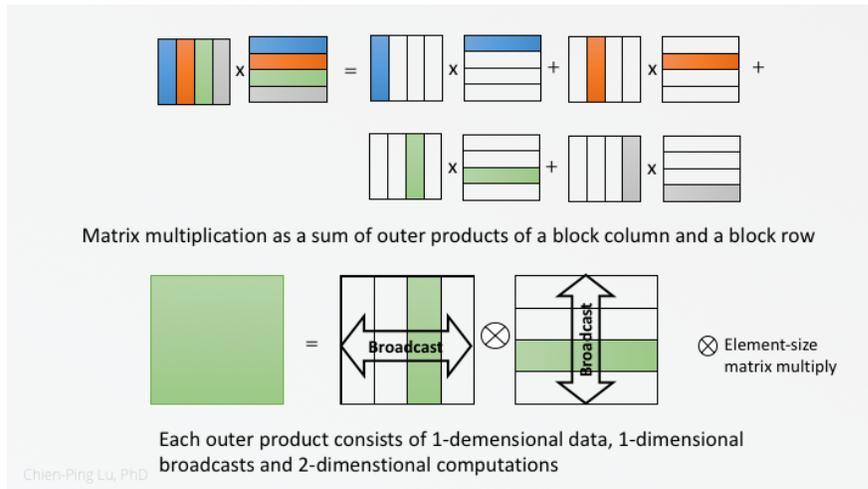
Figure 10 Matrix multiplication with outer products

The width of a block column and a block row can be a constant and is independent of the number of nodes. On a systolic array, the computations are also broken down into outer products. However, the width of the block column/row must match the side length of the systolic array to achieve optimal performance. Otherwise, the array is poorly occupied for problems with low inner dimension.

Outer product-based matrix multiplication algorithms, such as Scalable Universal Matrix Multiplication Algorithm (SUMMA) (Robert A. van de Geijn, 1995), have been proven to be very scalable both in theory and in practice in distributed systems.

**Efficient collective communications with logarithmic overhead**

The communication patterns in SUMMA or similar algorithms are based on collective communications defined for parallel computing on distributed systems. Advances in collective communication for HPC with recursive algorithms (Rajeev Thakur) reduce the communication overheads to be proportional to a logarithmic of the number of nodes and have been instrumental in the continuing performance growth in supercomputing.

---

**Native Supercomputing**

It is interesting to compare how matrix multiplication is achieved with a systolic array and a supercomputer, even though they are at completely different scales: one is on-chip and each node is a PE; the other is at the scale of a data center and each node is a compute cluster (Figure 11).

Broadcasts are implemented as forwarding data rightward, and reductions (a synonym of "accumulate" in the terminology of collective communications) are implemented as passing partial sums downward in a systolic array and accumulate along the way.

In comparison with an algorithm like SUMMA, broadcasts on a supercomputer happen in two dimensions among the nodes, while reductions are achieved in place at each node. There is no dependency, thus no dataflow but collective communication among the participating nodes. Since the reduction is in place, the number of nodes in either dimension is independent of the inner dimension of the matrices. As a matter of fact, the nodes don't even have to be arranged physically in a 2-dimensional topology as long as collection communication can be supported efficiently.

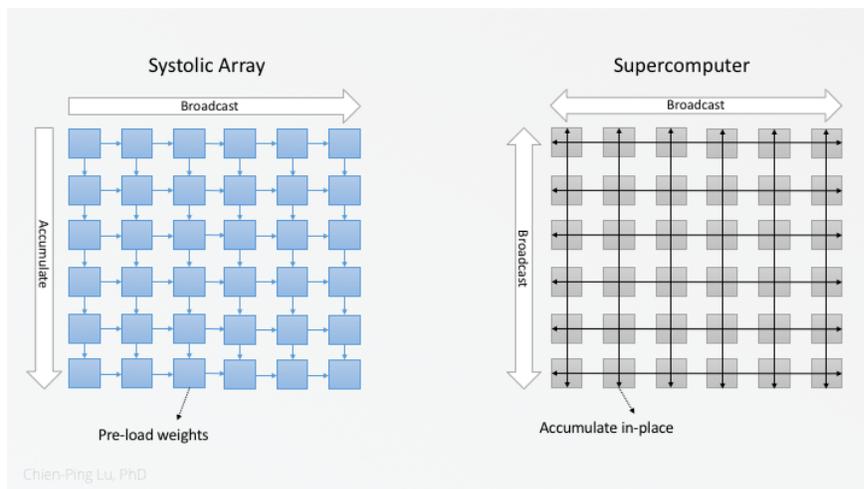

Figure 11 Matrix multiplication on a systolic array and a supercomputer

Today's distributed supercomputers are descendants of "Killer Micro" (Brooks, 1989), which were considered aliens invading the land of supercomputing in the early 90s. As a matter of fact, early supercomputers were purposely built to do matrix computations. Imagine that we build a supercomputer-on-chip by

1. Shrinking a compute cluster to a PE with only densely packed MAC units
2. Building on-chip data delivery fabric to support Collective Streaming, reminiscent of Collective Communication in Supercomputing

Just as efficient Collective Communication can be achieved recursively, efficient Collect Streaming can be accomplished recursively through the building block, Collective Streaming Element (CE). The CEs are inserted between the PEs and the memory to broadcast or scatter the data hierarchically to the PEs, and to reduce or gather the results recursively from the PEs. The 4 operations are analogous to the counterparts in collective communication in Supercomputing for the compute nodes to exchange data among themselves as shown in Figure 12. Compared to systolic arrays, the PEs do not have to be interlocked in a 2-dimensional grid and the latency can be within a constant factor of a logarithm of numbers of PEs. Building a supercomputer-on-chip can be considered as an effort to return to the matrix-centric root of supercomputing. It is effectively a *Native Supercomputer* (see Figure 13).

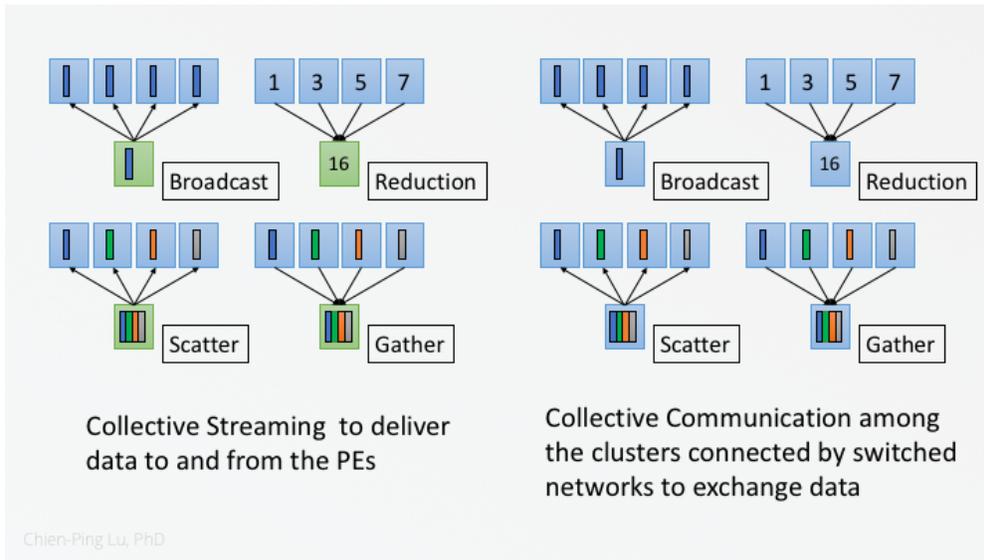

Figure 12 Collective Streaming vs. Collective Communication

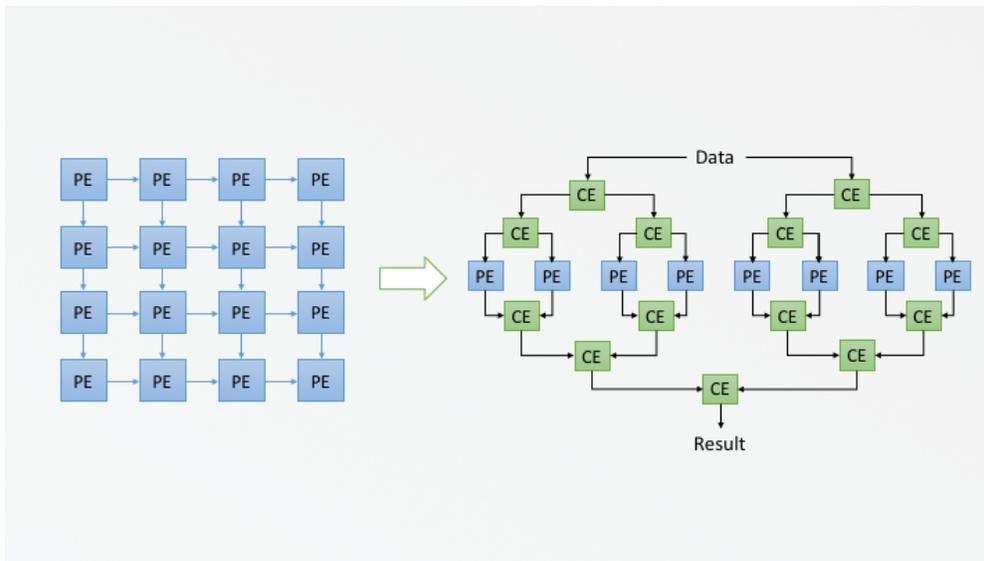

Figure 13 From a mesh to a hierarchically organized PEs

**Why Collective Streaming?**

In many *conventional parallel processors*, including the GPU, a core, as a universal computer, not only has to support many functions other than MAC, but also needs to retrieve data from the memory, expecting the data to be shared through memory hierarchy. As a result, it requires a significant investment in area and energy for generic functions, multiple levels of caches, scratch memory, and register files. Collective Streaming allows the computing units to comprise only MAC units without a deep memory hierarchy.

In a *spatial dataflow machine*, such as a systolic array, a PE still keep the legacy of a core having to communicate with other PEs. This causes latency and makes it difficult to scale. Collective Streaming allows orders of magnitude more MAC units without sacrificing latency.

A *programmable dataflow machine* is expected to resolve the dependencies among fine-grain data items. Given that dependencies among data items are collective, the efficiency of a programmable dataflow machine to handle generic data dependencies will be worse than a spatial dataflow machine.

---

**Conclusion**

As mentioned earlier, Turing envisioned a universal computer modeled after a human computer hired to do calculations with a pencil on paper. According to Turing, it will be the software that makes a machine thinks. However, it would be impossible for human programmers to write all the necessary software. Therefore, Turing envisioned a learning machine to generate the software. With Deep Learning, Turing's complete vision of the Computing Machinery for AI is fulfilled. It comprises a universal computer and a universal learning machine. His proposition of decoupling software from the underlying machinery remains intact. The role of a legacy universal computer will be like a CEO provisioning and formulating the time-consuming tasks into a universal learning machine, which does the heavy lifting.

AI was the inspiration behind the first Computing revolution. It shaped Computing as we know it today. Programmers were intrigued by the power of loops, subroutines, and recursions, and then learned to command them. The history of Computing now comes full circle. AI is coming back again to inspire Computing. Today, programmers are intrigued by the unreasonable effectiveness of Deep Neural Networks as Turing predicted:

> An important feature of a learning machine is that its teacher will often be very largely ignorant of quite what is going on inside, although he may still be able to some extent to predict his pupil's behaviour. (Turing, 1950)

Programmers will eventually command the new tools and drive the demand for the new computing machinery. The quest to make machines think will continue amid the slowdown of Moore's Law to drive the second Computing revolution. AI might not only maximize the remaining benefits of Moore's Law, but also revive Moore's Law beyond the current technology.